\documentclass[dvipsnames,format=sigconf,anonymous=false,review=false]{acmart} 

\def\vec#1{\mathchoice{\mbox{\boldmath$\displaystyle#1$}}
{\mbox{\boldmath$\textstyle#1$}}
{\mbox{\boldmath$\scriptstyle#1$}}
{\mbox{\boldmath$\scriptscriptstyle#1$}}}

\usepackage{booktabs} 
\usepackage{amsmath}
\usepackage{algorithm}
\usepackage[noend]{algpseudocode}
\usepackage[T1]{fontenc}
\usepackage{tabularx}
\usepackage{multirow}
\usepackage{array}
\usepackage{pbox}

\usepackage{color,soul}
\usepackage{nameref}

\usepackage{pgfplots}
\usepackage{graphicx}
\usepackage{subcaption}
\usepackage{tikz}
\usetikzlibrary{intersections, pgfplots.fillbetween}
\usepackage{siunitx}

\usepackage{url}
\usepackage{bm}

\definecolor{darkgreen}{rgb}{0.30, 0.50, 0.0}

\makeatletter
\renewcommand{\ALG@name}{Pseudocode}
\makeatother

\newcolumntype{L}[1]{>{\raggedright\let\newline\\\arraybackslash\hspace{0pt}}m{#1}}
\newcolumntype{C}[1]{>{\centering\let\newline\\\arraybackslash\hspace{0pt}}m{#1}}
\newcolumntype{R}[1]{>{\raggedleft\let\newline\\\arraybackslash\hspace{0pt}}m{#1}}

\newtheorem{thm}{Theorem}

\newcommand{\DSM}{D}

\newcommand{\VIG}{G}
\newcommand{\VIGnl}{G_{NL}}
\newcommand{\VIGnm}{G_{NM}}

\newcommand{\eVIG}{eG}

\graphicspath{{images/}}

\copyrightyear{2026}
\acmYear{2026}
\setcopyright{cc}
\setcctype{by-nc-nd}
\acmConference[GECCO '26]{Genetic and Evolutionary Computation Conference}{July 13--17, 2026}{San Jose, Costa Rica}
\acmBooktitle{Genetic and Evolutionary Computation Conference (GECCO '26), July 13--17, 2026, San Jose, Costa Rica}
\acmDOI{10.1145/3795095.3805172}
\acmISBN{979-8-4007-2487-9/2026/07}

\hypersetup{draft}
\begin{document}
    \title[Obtaining Partition Crossover masks using Statistical Linkage Learning]{Obtaining Partition Crossover masks using Statistical Linkage Learning for solving noised optimization problems with hidden variable dependency structure}

 	\author{Michal W. Przewozniczek}
\affiliation{
		\institution{Wroclaw Univ. of Science and Techn.}
		\city{Wroclaw}
		\country{Poland} 
	}
	\email{michal.przewozniczek@pwr.edu.pl}

 \author{Bartosz Frej}
 \affiliation{
 	\institution{Wroclaw Univ. of Science and Techn.}
 	\city{Wroclaw}
 	\country{Poland} 
 }
 \email{bartosz.frej@pwr.edu.pl}

 \author{Marcin M. Komarnicki}
 	\affiliation{
 		\institution{Wroclaw Univ. of Science and Techn.}
 		\city{Wroclaw}
 		\country{Poland} 
 	}
 	\email{marcin.komarnicki@pwr.edu.pl}

 \author{Michal Prusik}
 \affiliation{
 	\institution{Wroclaw Univ. of Science and Techn.}
 	\city{Wroclaw}
 	\country{Poland} 
 }
 \email{michal.prusik@pwr.edu.pl}

\author{Renato Tin\'os}
        \affiliation{
          \institution{University of S\~ao Paulo}
          \city{Ribeir\~ao Preto} 
          \country{Brazil}
        }
        \email{rtinos@ffclrp.usp.br}

	\renewcommand{\shortauthors}{Michal W. Przewozniczek et al.}

	\begin{abstract}
        In optimization problems, some variable subsets may have a joint non-linear or non-monotonical influence on the function value. Therefore, knowledge of variable dependencies may be crucial for effective optimization, and many state-of-the-art optimizers leverage it to improve performance. However, some real-world problem instances may be the subject of noise of various origins. In such a case, variable dependencies relevant to optimization may be hard or impossible to tell using dependency checks sufficient for problems without noise, making highly effective operators, e.g., Partition Crossover (PX), useless. Therefore, we use Statistical Linkage Learning (SLL) to decompose problems with noise and propose a new SLL-dedicated mask construction algorithm. We prove that if the quality of the SLL-based decomposition is sufficiently high, the proposed clustering algorithm yields masks equivalent to PX masks for the noise-free instances. The experiments show that the optimizer using the proposed mechanisms remains equally effective despite the noise level and outperforms state-of-the-art optimizers for the problems with high noise.
    \end{abstract}
	
	%
	%

	

	
	\maketitle
	

\section{Introduction}
\label{sec:intro}

Many state-of-the-art optimizers discover and utilize the knowledge about variable dependencies \cite{pxForBinary,ellGomea,FIHCwLL}. The quality of perturbation masks that are constructed concerning dependencies may be decisive for the overall optimizer performance \cite{linkageQuality}. Therefore, improving linkage-assisted variable clustering algorithms that yield such masks is an important research direction \cite{whitleyNext,ltgaOriginal,dsmga2,pxForBinary}. Constructing masks for problem instances subject to noise may become difficult. For problems that can be conveniently represented as a sum of subfunctions, variable dependency strength can be weighted \cite{GAwLL,verelSparseWalshModels}. This way, one can tell the most relevant dependencies from those that have only a negligible influence on the function value.\par

Some real-world problems can not be successfully decomposed with the non-linearity check \cite{GoldMonotonicity} and require the non-monotonicity check. If such problems are affected by noise, some dependencies may be significantly more relevant for the optimization than the other \cite{wVIG}. However, the weighted non-monotonicity check is not available. Therefore, we propose using Statistical Linkage Learning (SLL) \cite{ltga,P3Original} to optimize such problems. In contrast to non-linearity and non-monotonicity checks, SLL does not guarantee finding only the true dependencies. It aims at predicting variable dependency strength and, therefore, is applicable to decompose problems to which the aforementioned dependency checks do not apply.\par



Partition Crossover (PX) \cite{pxForBinary} is a dependency-based mixing operator that is effective in solving complex optimization problems but may become useless if irrelevant (e.g., noise-originated) dependencies occur. We propose a new SLL-dedicated variable clusterization algorithm that aims to produce masks that PX would yield if the problem were noise-free. To verify this proposition, we analyze the influence of noise on variable dependencies and propose mechanisms to introduce noise into the considered benchmark instances.

\section{Background}
\label{sec:relWork}

\subsection{Variable dependencies}
\label{sec:relWork:varDeps}

Any optimization pseudo-boolean function can be represented in the \textit{additive form} \cite{whitAdditivePolynomial}: $f(\vec{x})=\sum_{s=1}^{S} f_s(\vec{x}_{I_s})$,
where $\vec{x}=[x_1,\ldots,x_{n}]$ is a binary vector of size $n$, $I_s$ are subsets of $\{1,...,n\}$ ($I_s$ do not have to be disjoint), and $S$ is the number of these subsets.\par

If some or all $\vec{x}_{I_s}$ are disjoint, then discovering disjoint $I_s$ subsets allows processing (optimizing) the variables of each subfunction $f_s$ separately. Thus, such knowledge leads to a significant increase in the optimizer's efficiency and effectiveness \cite{ltga}. Intuitively, we can consider the variables within each $\vec{x}_{I_s}$ as dependent on each other.\par
If $\vec{x}_{I_s}$ overlap, then, in general, each $f_s$ cannot be processed separately. Nevertheless, perturbation masks based on variable dependencies remain decisive for the optimizers' effectiveness, also for such problems. For instance, these masks can be, efficiently used by transformation operators in Iterated Local Seach (ILS) \cite{ilsDLED} or serve as mixing masks for the crossover-like operators \cite{pxForBinary}. Therefore, variable dependencies are employed by state-of-the-art optimizers, and should be utilized whenever possible \cite{whitleyNext}. Among all, this recommendation applies to the \textit{k}-bounded problems, i.e., such that can be represented in the additive form, in which each $f_s$ takes no more than $k$ arguments \cite{transTokBounded}.\par

In black-box optimization, there are different strategies for discovering the interaction between variables (linkage learning). The non-linearity check allows discovering dependencies for problems for which the additive form is their convenient representation. Variables $x_g$ and $x_h$ are \textit{non-linearly} dependent if \cite{GoldMonotonicity} $f(\vec{x}) + f(\vec{x}^{g,h}) \neq f(\vec{x}^g) + f(\vec{x}^h)$,
where $\vec{x}^{g}$, $\vec{x}^h$, and $\vec{x}^{g,h}$ are the individuals obtained from $\vec{x}$ by flipping $x_g$, $x_h$ or both of them, respectively.\par

Consider: $f_{e1}(x_1,...,x_9) = f_{e1,1}(x_1,...,x_4) + f_{e1,2}(x_4,...,x_7) + \\ + f_{e1,3}(x_6,...,x_9)$, where each $f_{e1,i}(\vec{x_{I_i}}) = bim_4(u(\vec{x_{I_i}}))$, and $bim_k(u)$ is a bimodal deceptive function of order $k$ \cite{decBimodalOld}
\begin{equation}
    \label{eq:decBimod}
    \mathit{bim}_k\bigl(u(\vec{x})\bigr) = 
    \begin{cases}
        k / 2 - |u - k/2| - 1 & ,u \neq k \land u \neq 0\\
        k / 2 & ,u = k \lor u = 0\\
    \end{cases}
\end{equation}
    where $u$ is the sum of gene values (so-called \textit{unitation}), and $k$ is the deceptive function size (order). \par
\begin{table}  
    \caption{VIG for $f_{e1}$ } 
    \label{tab:Ex:fe1}
    \scriptsize
    \centering 
       \begin{tabular}{l|ccccccccc}
              & $x_1$  & $x_2$  & $x_3$ & $x_4$ & $x_5$ & $x_6$ & $x_7$ & $x_8$ & $x_9$\\
              \hline
            $x_1$ & X & \textbf{1} & \textbf{1} & \textbf{1} & 0 & 0 & 0 & 0 & 0 \\
            $x_2$ & \textbf{1} & X & \textbf{1} & \textbf{1} & 0 & 0 & 0 & 0 & 0 \\
            $x_3$ & \textbf{1} & \textbf{1} & X & \textbf{1} & 0 & 0 & 0 & 0 & 0 \\
            $x_4$ & \textbf{1} & \textbf{1} & \textbf{1} & X & \textbf{1} & \textbf{1} & \textbf{1} & 0 & 0 \\
            $x_5$ & 0 & 0 & 0 & \textbf{1} & X & \textbf{1} & \textbf{1} & 0 & 0 \\
            $x_6$ & 0 & 0 & 0 & \textbf{1} & \textbf{1} & X & \textbf{1} & \textbf{1} & \textbf{1} \\
            $x_7$ & 0 & 0 & 0 & \textbf{1} & \textbf{1} & \textbf{1} & X & \textbf{1} & \textbf{1} \\
            $x_8$ & 0 & 0 & 0 & 0 & 0 & \textbf{1} & \textbf{1} & X & \textbf{1} \\
            $x_9$ & 0 & 0 & 0 & 0 & 0 & \textbf{1} & \textbf{1} & \textbf{1} & X
        \end{tabular}
\end{table}

Note that $f_{e1}$ is an overlapping function and each subfunction may be found hard to solve. All variables that are the entry arguments of each subfunction are mutually non-linearly dependent on each other. The network of these dependencies can be represented by the Variable Interaction Graph $\VIG$ (VIG). $\VIG$ is an undirected graph in which each node refers to a given variable and if $x_g$ and $x_h$ are dependent (e.g., non-linearly), then the nodes are joined by an edge, i.e., $\VIG(g,h)=\VIG(h,g)=1$. Frequently, VIGs concern non-linear dependencies (e.g., \cite{whitleyNext,pxForBinary}). $\VIG$ can be represented by a matrix. The VIG of $f_{e1}$ is presented in Table \ref{tab:Ex:fe1}.\par



In the gray-box optimization setting, where $\VIG$ is known \textit{a priori}, frequently, Walsh decomposition is employed to discover all non-linear dependencies \cite{heckendorn2002,GrayBoxWhitley,superMasksGray}. For many problems, non-linear dependencies, the additive form and Walsh decomposition are \textbf{\textit{not}} convenient. Consider a trivial problem (denoted as \textit{onemax}) $f_{e2}(x_1,...,x_n) = u(x_1,...,x_n)$. Any local search procedure that tries to flip bits in any order and preserves the improving modifications is guaranteed to find the optimal solution of $f_{e2}$ in $n+1$ fitness function evaluations (FFE). Indeed, $\VIGnl$ of $f_{e2}$ ($\VIG$ concerning non-linear dependencies) is an empty graph. Thus, each variable can be optimized separately. Note that the aforementioned local search algorithm that solves $f_{e2}$ will be equally effective and efficient for $f_{e3}(x_1,...,x_n) = (f_{e2}(x_1,...,x_n))^2$ because, in $f_{e3}$, each variable can be optimized separately, too. However, for $f_{e3}$, $\VIGnl$ will be a full graph, which is a (false) premise to process all $f_{e3}$ variables jointly.\par

The non-monotonicity check was proposed to overcome the oversensitivity of the non-linearity check and ignore the easy non-linear dependencies shown in the example above \cite{GoldMonotonicity}. We consider its more sensitive version for which $x_g$ and $x_h$ are \textit{non-monotonically} dependent if at least one of the following clauses holds \cite{2dled}.\\
\footnotesize
    \textbf{C1.} $f(\vec{x}) < f(\vec{x}^g)  \land f(\vec{x}^h) \geq f(\vec{x}^{g,h})$   \textbf{C4.} $f(\vec{x}) < f(\vec{x}^h)  \land f(\vec{x}^g) \geq f(\vec{x}^{g,h})$\\
    \textbf{C2.} $f(\vec{x}) = f(\vec{x}^g)  \land f(\vec{x}^h) \neq f(\vec{x}^{g,h})$    \textbf{C5.} $f(\vec{x}) = f(\vec{x}^h)  \land f(\vec{x}^g) \neq f(\vec{x}^{g,h})$\\
    \textbf{C3.} $f(\vec{x}) > f(\vec{x}^g)  \land f(\vec{x}^h) \leq f(\vec{x}^{g,h})$   \textbf{C6.} $f(\vec{x}) > f(\vec{x}^h)  \land f(\vec{x}^g) \leq f(\vec{x}^{g,h})$\\
\normalsize
For $f_{e3}$, $\VIGnm$ ($\VIG$ concerning the non-monotonicity check) is an empty graph. Note that the additive form is not suitable to represent $f_{e3}$. Consider $f_{e4}(x_1,...,x_9) = f_{e4,1}(x_1,...,x_4) \cdot f_{e4,2}(x_4,...,x_7) \cdot \\ f_{e4,3}(x_6,...,x_9)$, where each $f_{e4,i} = bim_4(\vec{x}_{I_i}) + 1$. For $f_{e4}$, $\VIGnl$ is a full graph (which makes it useless), but $\VIGnm$ equals $\VIG$ presented in Table \ref{tab:Ex:fe1}. For each $f_{e4,i}$, all of its variables depend on each other. \par

\subsection{Partition Crossover and its features}
\label{sec:relWork:PX}

\begin{table}  
    \caption{Remaining VIG for the PX mask creation example } 
    \label{tab:Ex:PX}
    \scriptsize
    \centering 
       \begin{tabular}{l|ccccccccc}
              & $x_1$  & $x_2$  & $x_3$ & $x_4$ & $x_5$ & $x_6$ & $x_7$ & $x_8$ & $x_9$\\
              \hline
            $x_1$ & X & 0 & 0 & 0 & 0 & 0 & 0 & 0 & 0 \\
            $x_2$ & 0 & X & \textcolor{red}{\textbf{1}} & 0 & 0 & 0 & 0 & 0 & 0 \\
            $x_3$ & 0 & \textcolor{red}{\textbf{1}} & X & 0 & 0 & 0 & 0 & 0 & 0 \\
            $x_4$ & 0 & 0 & 0 & X & 0 & 0 & 0 & 0 & 0 \\
            $x_5$ & 0 & 0 & 0 & 0 & X & \textcolor{blue}{\textbf{1}} & 0 & 0 & 0 \\
            $x_6$ & 0 & 0 & 0 & 0 & \textcolor{blue}{\textbf{1}} & X & 0 & \textcolor{blue}{\textbf{1}} & 0 \\
            $x_7$ & 0 & 0 & 0 & 0 & 0 & 0 & X & 0 & 0 \\
            $x_8$ & 0 & 0 & 0 & 0 & 0 & \textcolor{blue}{\textbf{1}} & 0 & X & 0 \\
            $x_9$ & 0 & 0 & 0 & 0 & 0 & 0 & 0 & 0 & X
        \end{tabular}
\end{table}

Partition Crossover (PX) was proposed to utilize information stored in $\VIG$ \cite{pxForBinary}. PX creates a dedicated set of masks for each pair of mixed individuals, say $\vec{x_1}$ and $\vec{x_2}$, using $\VIG$. It removes all edges referring to at least one variable that is equal in $\vec{x_1}$ and $\vec{x_2}$, i.e., a \emph{PX mask} is a connected component in such a graph, which consists only of the vertices at which $\vec{x_1}$ and $\vec{x_2}$ differ. Consider $\VIG$ presented in Table \ref{tab:Ex:fe1} and two solutions for $f_{e1}$:\\
$\vec{x_1} = [1111\text{ }001\text{ }01]$; $\vec{x_{1,I_1}} = [1111]$ $\vec{x_{1,I_2}} = [1001]$ $\vec{x_{1,I_3}} = [0101]$\\
$\vec{x_2} = [1001\text{ }111\text{ }11]$; $\vec{x_{2,I_1}} = [1001]$ $\vec{x_{2,I_2}} = [1111]$ $\vec{x_{2,I_3}} = [1111]$\\
Thus, all edges that refer to variables $x_1$, $x_4$, $x_7$, and $x_9$ will be ignored (removed from $\VIG$) during PX masks creation. For the remaining $\VIG$, PX will cluster all variables that have at least one common edge. Table \ref{tab:Ex:PX} presents the remaining $\VIG$ for $\vec{x_1}$ and $\vec{x_2}$ mixing. The set of PX masks for these individuals contains two PX masks (\textcolor{red}{\textbf{red}} $=\{x_2, x_3\}$ and \textcolor{blue}{\textbf{blue}} $=\{x_5,x_6,x_8\}$). If we use the blue mask, we will obtain the offspring individuals $x_{o1}$ and $x_{o2}$, where $x_{o1}$ is a copy of $\vec{x_1}$ with the genes from $\vec{x_2}$ marked by the blue mask, while $x_{o2}$ is created analogously.\\
$\vec{x_{o1}} = [1111\textcolor{blue}{\textbf{1}}\textcolor{blue}{\textbf{1}}1\textcolor{blue}{\textbf{1}}1]$; $\vec{x_{o1,I_1}} = [1111]$  $\vec{x_{o1,I_2}} = [1\textcolor{blue}{\textbf{1}}\textcolor{blue}{\textbf{1}}1]$ $\vec{x_{o1,I_3}} = [\textcolor{blue}{\textbf{1}}1\textcolor{blue}{\textbf{1}}1]$\\
$\vec{x_{o2}} = [1001\textcolor{blue}{\textbf{0}}\textcolor{blue}{\textbf{0}}1\textcolor{blue}{\textbf{0}}1]$; $\vec{x_{o2,I_1}} = [1001]$ $\vec{x_{o2,I_2}} = [1\textcolor{blue}{\textbf{0}}\textcolor{blue}{\textbf{0}}1]$ $\vec{x_{o2,I_3}} = [\textcolor{blue}{\textbf{0}}1\textcolor{blue}{\textbf{0}}1]$\par

Note that in $x_{o1}$ and $x_{o2}$ the variable value strings that refer to each subfunction $f_{e1,i}$ where not broken by the mixing despite the overlaps. For instance, for $\vec{x_{I_2}}$ strings $1111$ and $1001$ occur exactly once in $\vec{x_1}$ and $\vec{x_2}$ and exactly once in $x_{o1}$ and $x_{o2}$.

PX is proven to preserve complete subfunction argument sets \cite{dgga}. This feature has important consequences. If $\VIG$ concerns non-linear dependencies and the subfunctions are added, then $f(\vec{x_1}) + f(\vec{x_2}) = f(\vec{x_{1,I_1}}) + ... + f(\vec{x_{1,I_S}}) + f(\vec{x_{2,I_1}}) + ... + f(\vec{x_{2,I_S}})$. Since all $\vec{x_{I_i}}$ that differ in $\vec{x_1}$ and $\vec{x_2}$ occur exactly once in $\vec{x_1}$ and $\vec{x_2}$ and exactly once in $\vec{x_{o1}}$ and $\vec{x_{o2}}$, then $f(\vec{x_{o1}}) + f(\vec{x_{o2}}) = f(\vec{x_1}) + f(\vec{x_2})$ because each added component occurs the same number of times on both sides of the equation.\par

For $\VIGnm$, PX also preserves the subfunction argument value sets. For instance, for $f_{e4}$, PX will preserve the argument value sets of all $f_{e4,i}$. In that case, one of the following fitness relations between parent and offspring individuals will be true \cite{dgga} (we refer to the same solutions as in the above PX example concerning $f_{e1}$):\\
\textcolor{white}{oo}If $f(\vec{x_1}) < f(\vec{x_{o1}})$ then $f(\vec{x_2}) > f(\vec{x_{o2}})$\\
\textcolor{white}{oo}If $f(\vec{x_1}) = f(\vec{x_{o1}})$ then $f(\vec{x_2}) = f(\vec{x_{o2}})$\\
\textcolor{white}{oo}If $f(\vec{x_1}) > f(\vec{x_{o1}})$ then $f(\vec{x_2}) < f(\vec{x_{o2}})$\\
PX using $\VIGnl$ or $\VIGnm$ is guaranteed to yield offspring that improve one of the mixed individuals and deteriorate the other, or the fitness of each offspring is equal to the fitness of the appropriate parent. Therefore, PX may can significantly improve the optimizer's effectiveness when solving overlapping problems \cite{pxForBinary}. Initially, PX was dedicated to the gray-box optimization settings for which all dependencies are known \textit{a priori}. In gray-box optimization, when the additive structure of the objective function is known, and the values of the subfunctions can be evaluated for every combination of their partial solutions, PX \cite{pxForBinary} explicitly exploits this structural information. Given this knowledge, it is possible to greedily select the best parent for each mask. Then, PX finds the best of $2^M$ possible offspring, where $M$ is the number of masks.\par

Recently, PX was employed for the black-box scenarios that use \textit{empirical} $\VIG$ ($\eVIG$) \cite{dgga}, i.e., containing only the true dependencies but potentially incomplete. In addition, since partial evaluations are not available, subsets of variables (partial solutions) defined by the masks are recombined one at a time. While PX for black-box optimization does not permit the parallel evaluation of multiple offspring, in contrast to PX for gray-box optimization, it is applicable to any problem. An important property of PX is that recombining local optima may yield a local (or even global) optimum. By recombining partial solutions corresponding to one mask at a time, PX for black-box optimization generates offspring, whether local optima or not, that remain close to their parents. This property is particularly appealing when PX is used within iterative frameworks that repeatedly generate local optima, such as the approach proposed here. In the black-box scenario, $\eVIG$ is constructed iteratively: it initially contains no edges, and new edges are added whenever an interaction is discovered. In such a case, PX may be used to detect missing dependencies in the $\eVIG$ it uses. More information about missing linkage detection can be found in \cite{dgga}.\par

PX allows solving complex overlapping problems. However, its capability is limited by the \textit{epistasis} of $\VIG$ it uses, i.e., the percentage of pairs marked as dependent. If epistasis is too high, then PX will frequently (or always) yield masks that join all genes that differ in the mixed individuals. Therefore, some researchers propose weighting dependencies and considering only the strongest ones when constructing mixing masks \cite{GAwLL,wVIG}. Another proposal is to leverage certain features of the Walsh-based function model to split clusters of dependent variables into shorter masks while preserving the PX features \cite{superMasksGray}. However, none of these proposals is suitable for problems where the non-linearity check is not convenient, and considering non-monotonical dependencies is a must.\par

\subsection{Statistical linkage learning -- operators}
\label{sec:relWork:sll}

Statistical Linkage Learning (SLL) differs significantly from the direct dependency checks presented in Section \ref{sec:relWork:varDeps}. SLL techniques use statistical analysis to predict the existence of dependencies rather than discover them \cite{3lo}. Thus, SLL outputs the predicted dependency strength that arises from the statistics it uses. SLL can not tell direct and indirect dependencies (e.g., $x_1$ in $f_{e1}$ is directly dependent on $x_2$, $x_3$, and $x_4$, and indirectly dependent on the rest of the variables). Nevertheless, SLL was formally proven to be highly efficient and precise for decomposing some problems \cite{linkageQuality}, as well as being inefficient and imprecise in decomposing some others \cite{sllForBimodals}. Nevertheless, its overall precision and efficiency are high enough to be a part of many state-of-the-art optimizers \cite{ltgaPopulationSizing,P3Original}. 

Instead of constructing a $\VIG$, SLL builds the Dependency Structure Matrix $\DSM$ (DSM). To this end, it analyzes the frequencies of variable pair values that occur in the population. On this basis, it estimates the dependency strength for a given variable pair. Mutual information is one of the frequently employed measures \cite{mutualInformation} $I(X, Y) = \sum_{x \in \mathcal{X}} \sum_{y \in \mathcal{Y}} p_{X,Y}(x, y) \log_2 \frac{p_{X,Y}(x,y)}{p_X(x)p_Y(y)}$
    where $X$ and $Y$ are two variables considered random variables with values in $\mathcal{X}$ and $\mathcal{Y}$ and distribution functions $p_X$ and $p_Y$, respectively, while $p_{X,Y}$ is their joint distribution. The entries of $\DSM$ are given by the normalized information: $D(X,Y) = \frac{I(X,Y)}{H(X,Y)}$, where $H(X,Y)$ is the joint entropy of the variables $X$ and $Y$. Note that often a metric given by $1-D(X,Y)$ is used instead \cite{dsmga2}. \par

    Similarly to VIGs, each $\DSM$ entry $\DSM(g,h) = D(h,g)$ refers to $x_g$ and $x_h$, but informs about the dependency strength, not its objective existence. Linkage Trees (LTs) are frequently used to cluster DSMs (see Section S-I, supplementary material, for the LT construction algorithm). LT nodes are useful perturbation masks. Among all, they are employed in Optimal Mixing (OM) \cite{substructures}. OM aims to improve the \textit{source} individual $\vec{x_{src}}$ by using LT nodes as mixing masks. Some optimizers, e.g., LT Gene-pool Optimal Mixing Evolutionary Algorithm (LT-GOMEA) \cite{ltgaPopulationSizing}, consider LT nodes in a random order. The others, e.g., Parameter-less Population Pyramid (P3) \cite{P3Original}, use random ordering within nodes of the same size, and consider the shorter nodes first. OM in P3 ignores all nodes of size $1$.\par
    
    For each $mask$, OM randomly chooses a \textit{donor} individual $\vec{x_{don}}$ from the population such that there exists at least one variable covered by the $mask$ that differs in $\vec{x_{don}}$ and $\vec{x_{src}}$.    
    Using mask, OM creates the candidate solution $\vec{x'_{src}} = \vec{x_{src}} \xleftarrow[mask]{} \vec{x_{don}}$. If $f(\vec{x'_{src}}) \geq f(\vec{x_{src}})$, then $\vec{x'_{src}}$ replaces $\vec{x_{src}}$. 
    

    SLL-using optimizers (LT-GOMEA and P3), and their versions employing the First Improvement Hill Climber with Linkage Learning (FIHCwLL) \cite{FIHCwLL} (supporting low-cost non-monotonical dependency discovery), are presented in the supplementary material.

\subsection{Noised problems with the hidden structure}
\label{sec:relWork:nooise}

Noise may influence the value of the optimized function for many real-world problems, e.g, in feature selection problems, we evaluate feature choices by training a classifier and evaluating the quality of its output. Thus, each evaluation of the same feature choice may yield a slightly different evaluation value. 
Intuitively, we can define a \textit{noised problem} as $f_{noised}(\vec{x}) = f_{true}(\vec{x}) + f_{noise}(\vec{x})$
where $f_{true}(\vec{x})$ is the optimized function and $f_{noise}(\vec{x})$ disturbs the $f_{true}(\vec{x})$ value.

Thus, for the noised problem, focusing on optimizing $f_{true}(\vec{x})$ seems easy if the influence of $f_{noise}(\vec{x})$ on the optimized function value is negligible for most of the solutions of the search space. However, the noised problems are frequently characterized by VIGs that are full graphs, because even a very low noise volume may raise dependencies between all variable pairs \cite{wVIG}. Full VIGs are useless for PX. Thus, to solve a given noised problem effectively, we must discover only the dependencies that arise from $f_{true}(\vec{x})$ and ignore the dependencies arising from $f_{noise}(\vec{x})$. Therefore, the \textit{dependency structure} of $f_{true}(\vec{x})$ can be considered \textit{hidden}.\par

To handle the noised problems, various weighted dependency checks were proposed in \cite{wVIG,GAwLL}. However, these checks apply only to problems for which non-linear dependencies are convenient for $f_{true}(\vec{x})$. For the noised real-world problems, such that $f_{true}(\vec{x})$ requires the use of non-monotonical dependencies, the aforementioned checks are not suitable.\par

Many instances of noised real-world problems may require non-monotonical dependencies \cite{wVIG}. This dependency type was also shown to be significantly more resistant to discovering dependencies arising from $f_{noise}(\vec{x})$. Reformulating the optimized function to decrease the noise or make non-linear dependencies convenient for $f_{true}(\vec{x})$ is frequently impossible. In the feature selection problem presented at the beginning of this subsection, we know the exact function that evaluates the classifier's performance. However, even if the classifier training algorithm is deterministic, the function that \textit{directly} transforms the chosen feature set into its evaluation may remain unknown. The same holds for a wide class of real-world problems that use an algorithm to construct the solution from the encoded genotype before evaluating it, e.g., timetabling \cite{automated_timetabling}, multi-echelon supply chains \cite{multi-echelon}.\par

To summarize, many complex real-world problems can be considered overlapping \cite{conflictingOverlapsAreRealWorld} and noised problems which require using the non-monotonicity check \cite{wVIG}. 
For such problems, P3-FIHCwLL and LT-GOMEA-FIHCwLL (P3 and LT-GOMEA using direct checks to support SLL) \cite{FIHCwLL} may lose their advantage over their original versions because $\eVIG$ they use will quickly become a full (or close to full) graph. Therefore, we propose a new way of constructing LTs using SLL that applies to any search space. We prove that if the DSM quality is sufficiently high, our proposal is guaranteed to yield an LT that contains a full set of PX masks.

\section{PX-like Linkage Trees}
\label{sec:pxLTs}


\subsection{Linkage Trees disadvantages}
\label{sec:pxLTs:LTweaknesses}

The LT creation algorithm presented in Section S-I (supplementary material) joins the most dependent variable pairs first. Such a procedure seems intuitive and justified. However, for overlapping problems, it may yield surprising and unfavorable mixing masks. Let us define a \textit{perfect} DSM.
\newcommand{\thr}{\theta}
\textcolor{black}{
\begin{definition} \label{def:perfDSM}
    For a given $f(x)$ with VIG $\VIG$, a DSM $\DSM^*$ is \emph{perfect} if  $\VIG(g,h) = 1$ and $\VIG(i,j) = 0$ implies  $\DSM^*(g,h) > \DSM^*(i,j)$ for any choice of  $g$, $h$, $i$, and $j$.
\end{definition}
We emphasize the difference in our usage of VIG and DSM: for a given problem and assumed dependency type (non-monotonic, non-linear) the VIG is a theoretical object uniquely assigned to the problem; a DSM is a matrix obtained experimentally by means of SLL, which guesses the 
stochastic dependencies.
A perfect DSM is a matrix that allows recovering VIG $\VIG$ with the following rule: for some threshold $\thr$ \textbf{if} $\DSM(g,h) > \thr$ \textbf{then} $\VIG(g,h) = 1$ \textbf{else} $\VIG(g,h) = 0$. Note that though the number of possible perfect DSMs is infinite, in practice they may be unattainable, if the stochastic behavior in the optimization process poorly reflects dependencies described by VIG (see examples in \cite{sllForBimodals}). Moreover, the correct value of the threshold $\thr$ cannot be obtained from the DSM, without the knowledge of VIG, but its existence is guaranteed with the above definition.}\par

Throughout the current section, $D^*$ will denote a \emph{perfect} DSM. Since in $\DSM^*$ all true dependencies (such that $\VIG(g,h)=1$) are assigned a higher value of the dependency strength than any false dependency (such that $\VIG(g,h)=0$), then we may expect that an LT built on the basis of $D^*$ will always contain useful mixing masks.

\begin{table} 
    \caption{Exemplary $\DSM^*$ for $f_{e1}$ } 
    \label{tab:Ex:DSfe1}
    \scriptsize
    \centering 
       \begin{tabular}{l|ccccccccc}
              & $x_1$  & $x_2$  & $x_3$ & $x_4$ & $x_5$ & $x_6$ & $x_7$ & $x_8$ & $x_9$\\
              \hline
            $x_1$ & X & \textcolor{red}{0.99} & \textcolor{red}{0.51} & \textcolor{red}{0.51} & \textcolor{blue}{0.25} & \textcolor{blue}{0.25} & \textcolor{blue}{0.25} & \textcolor{blue}{0.25} & \textcolor{blue}{0.25} \\
            $x_2$ & \textcolor{red}{0.99} & X & \textcolor{red}{0.51} & \textcolor{red}{0.51} & \textcolor{blue}{0.25} & \textcolor{blue}{0.25} & \textcolor{blue}{0.25} & \textcolor{blue}{0.25} & \textcolor{blue}{0.25} \\
            $x_3$ & \textcolor{red}{0.51} & \textcolor{red}{0.51} & X & \textcolor{red}{0.99} & \textcolor{blue}{0.49} & \textcolor{blue}{0.50} & \textcolor{blue}{0.50} & \textcolor{blue}{0.50} & \textcolor{blue}{0.50} \\
            $x_4$ & \textcolor{red}{0.51} & \textcolor{red}{0.51} & \textcolor{red}{0.99} & X & \textcolor{red}{0.60} & \textcolor{red}{0.98} & \textcolor{red}{0.73} & \textcolor{blue}{0.50} & \textcolor{blue}{0.50} \\
            $x_5$ & \textcolor{blue}{0.25} & \textcolor{blue}{0.25} & \textcolor{blue}{0.49} & \textcolor{red}{0.60} & X & \textcolor{red}{0.60} & \textcolor{red}{0.60} & \textcolor{blue}{0.49} & \textcolor{blue}{0.50} \\
            $x_6$ & \textcolor{blue}{0.25} & \textcolor{blue}{0.25} & \textcolor{blue}{0.50} & \textcolor{red}{0.98} & \textcolor{red}{0.60} & X & \textcolor{red}{0.73} & \textcolor{red}{0.73} & \textcolor{red}{0.55} \\
            $x_7$ & \textcolor{blue}{0.25} & \textcolor{blue}{0.25} & \textcolor{blue}{0.50} & \textcolor{red}{0.73} & \textcolor{red}{0.60} & \textcolor{red}{0.73} & X & \textcolor{red}{0.99} & \textcolor{red}{0.55} \\
            $x_8$ & \textcolor{blue}{0.25} & \textcolor{blue}{0.25} & \textcolor{blue}{0.50} & \textcolor{blue}{0.25} & \textcolor{blue}{0.49} & \textcolor{red}{0.73} & \textcolor{red}{0.99} & X & \textcolor{red}{0.55} \\
            $x_9$ & \textcolor{blue}{0.25} & \textcolor{blue}{0.25} & \textcolor{blue}{0.50} & \textcolor{blue}{0.50} & \textcolor{blue}{0.50} & \textcolor{red}{0.55} & \textcolor{red}{0.55} & \textcolor{red}{0.55} & X
        \end{tabular}
\end{table}

\begin{figure}[]
    \begin{subfigure}[b]{0.45\linewidth}
		\resizebox{\linewidth}{!}{%
			\centering
	       \includegraphics[width=0.35 \linewidth]{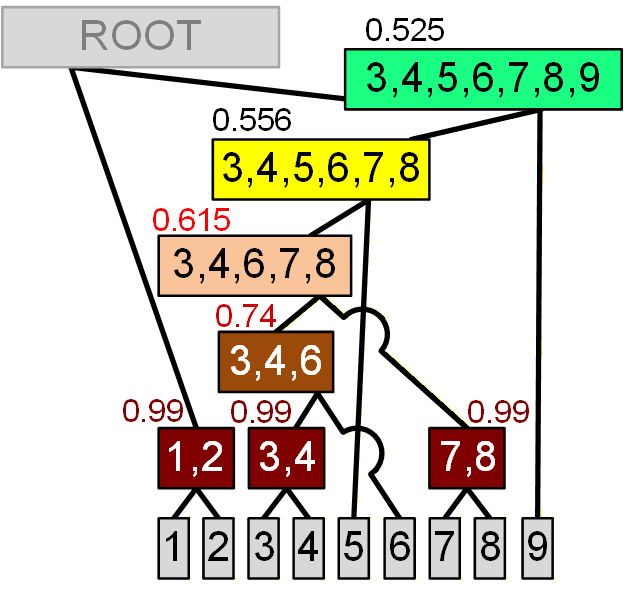}
	
		}
		\caption{Standard LT}
	   \label{fig:LTs:classic}
	\end{subfigure}
    \begin{subfigure}[b]{0.4\linewidth}
        \resizebox{\linewidth}{!}{%
			\centering
	       \includegraphics[width=0.4 \linewidth]{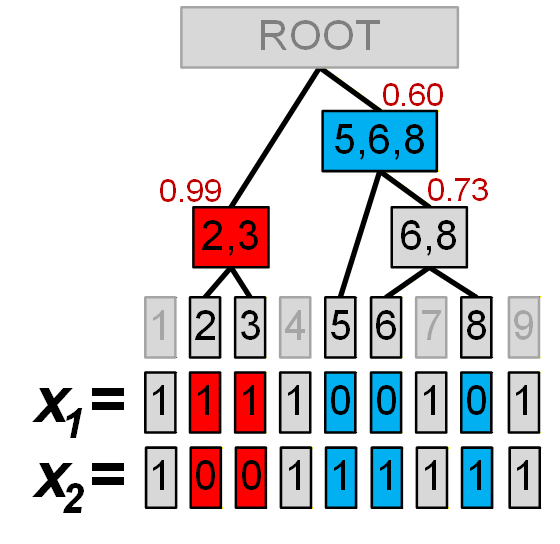}
	
		}
		
	   \caption{PX-LT}
	   \label{fig:LTs:pxLT}
	\end{subfigure}
	
	\caption{LTs for $\DSM^*$ in Table \ref{tab:Ex:DSfe1}}
	\label{fig:LTs}
\end{figure}

Consider $f_{e1}$ and its exemplary $\DSM^*$ presented in Table \ref{tab:Ex:DSfe1}. All non-monotonical dependencies are marked in red, and their DSM-based strength is higher than $0.51$. The dependency strength between variable pairs that are not dependent is marked in blue and their values are not higher than $0.50$. Fig. \ref{fig:LTs:classic} presents an LT that would be created for the considered $\DSM^*$ (number at each node reports the dependency strength that will cause the creation of a given node).\par

First, nodes $\{1,2\}$, $\{3,4\}$, and $\{7,8\}$ are joined because the dependency strength for these variable pairs is the highest ($0.99$). The next created node is $\{3,4,6\}$ because its dependency strength equals $0.74$ and is the highest for all remaining node pairs ($(\DSM^*(3,6) + \DSM^*(4,6))/2 = 0.74$). The rest of LT nodes is created analogously.\par

Consider $\vec{x_1}$ and $\vec{x_2}$, the solutions to $f_{e1}$ presented in Section \ref{sec:relWork:PX}. Mixing these solutions using PX masks can improve them, i.e., for the \textcolor{red}{\textbf{red}} mask, $f(\vec{x_{o2}}) > f(\vec{x_{2}})$, while for the \textcolor{blue}{\textbf{blue}} mask, $f(\vec{x_{o1}}) > f(\vec{x_{1}})$. However, none of the LT nodes presented in Fig. \ref{fig:LTs:classic} will yield improvement for mixing $\vec{x_1}$ and $\vec{x_2}$. Moreover, despite building LT on the basis of $\DSM^*$ that seems to be a DSM of a high quality, except for the root, none of the masks covers the complete set of differences for a given set of subfunctions.\par

\subsection{PX-like Linkage Trees}
\label{sec:pxLTs:pxLTalg}

The example in the previous section shows that it may be hard to yield high-quality mixing masks using LTs even when having a high-quality DSM. This observation is confirmed by the prior results indicating that to solve overlapping problems by LT-using optimizers, it is favorable to maintain many different LTs \cite{3lo}. Therefore, we propose PX-like Linkage Trees (PX-LTs), a new DSM clusterization that is guaranteed to yield PX-like mixing masks if it is using any $\DSM^*$. Thus, for a high-quality DSMs, PX-LTs should significantly increase the effectiveness of the SLL-using optimizers for those problems for which PX is an effective mixing operator. \par

\begin{algorithm}
    \small
	\caption{PX-like Linkage Tree construction algorithm}
	\begin{algorithmic}[1]
        \Function{ConstructLT}{$\DSM$, $\vec{x_{p1}}$, $\vec{x_{p2}}$}
            \State  $curNodes \gets$ createSingleNodeForEachVariable(); \label{alg:pxLTcreation:nodeVarStart}
            \For{$i$ \textbf{in} $\{1,...,n\}$}
                \If {$\vec{x_{p1}}[i] = \vec{x_{p2}}[i]$} 
                    \State $curNodes \gets $ RemNode($curNodes$,$x_i$); \label{alg:pxLTcreation:nodeVarEnd}
                \EndIf
            \EndFor

            \State  $depStrengths \gets $ empty;\label{alg:pxLTcreation:depsInitStart}
            \For{\textbf{each} $n1$ \textbf{in} $curNodes$}
                \For{\textbf{each} $n2 \neq n1$ \textbf{in} $curNodes$}
                    \State $depStrengths \gets $ AddCell($n1$,$n2$, $\DSM[n1][n2]$);\label{alg:pxLTcreation:depsInitEnd}
                \EndFor
            \EndFor
            
            \State  $allNodes \gets curNodes$;

            \While{$|curNodes| > 1$}
                \State $node1, node2 \gets$ GetNodesHighestDependency($depStrengths$);
                \State $depStrengths \gets $ Remove($depStrengths$, $node1$);
                \State $depStrengths \gets $ Remove($depStrengths$, $node2$);
                
                \State $curNodes \gets curNodes \setminus node1$;
                \State $curNodes \gets curNodes \setminus node2$;
                \State $newNode \gets node1 \cup node2$

                \For{\textbf{each} $node$ \textbf{in} $curNodes$}\label{alg:pxLTcreation:newStrStart}
                    \State $newStr \gets 0$;
                    \For{\textbf{each} $varN$ \textbf{in} $newNode$}
                        \For{\textbf{each} $varO$ \textbf{in} $node$}
                            \If {$newStr < \DSM[varN][varO]$}
                                \State $newStr \gets \DSM[varN, varO]$;
                            \EndIf
                        \EndFor
                    \EndFor
                    \State $depStrength \gets $ AddCell($newNode$,$node$, $newStr$);\label{alg:pxLTcreation:newStrEnd}
                \EndFor

                \State $curNodes \gets curNodes + newNode$;
                \State $allNodes \gets allNodes + newNode$;
            \EndWhile
            \State \Return{$allNodes$};
        \EndFunction
    \end{algorithmic}
	\label{alg:pxLTcreation}
\end{algorithm}

In the example from the prior section, variables $x_3$ and $x_6$ would not be joined in the same PX mask. In LT, they are joined in the node $\{x_3, x_4, x_6\}$ because they both depend on $x_4$, which equals in $\vec{x_1}$, $\vec{x_2}$, and is irrelevant for mixing these individuals. Therefore, in the PX-LT construction algorithm, we ignore all variable dependencies that refer to at least one variable that is equal in the mixed individuals.\par


The original LT construction algorithm uses the average DSM-strength for nodes that group more than one variable. PX joins dependent variables. In $\DSM^*$, the higher values refer to those pairs that are dependent in $\VIG$. Therefore, to obtain PX masks from $\DSM^*$, we wish to join those nodes whose variables are the most dependent on each other in $\DSM^*$. In the proposed PX-LT construction algorithm, the dependency strength for nodes grouping more than one variable uses the maximal $\DSM$ value for variable pairs that can be created between two nodes instead of using an average.\par

Pseudocode \ref{alg:pxLTcreation} presents the proposed PX-LT construction algorithm for $\DSM$ and two parent individuals ($\vec{x_{p1}}$ and $\vec{x_{p2}}$). First, we create nodes for all variables that differ in $\vec{x_{p1}}$ and $\vec{x_{p2}}$ (lines \ref{alg:pxLTcreation:nodeVarStart}-\ref{alg:pxLTcreation:nodeVarEnd}) and initialize the matrix of their dependencies using $\DSM$ (lines \ref{alg:pxLTcreation:depsInitStart}-\ref{alg:pxLTcreation:depsInitEnd}). The rest of the procedure is the same as for the regular LT creation, except for using the maximal $\DSM$ value for the dependency strength between the two nodes instead of using the average (lines \ref{alg:pxLTcreation:newStrStart}-\ref{alg:pxLTcreation:newStrEnd}).\par

Fig. \ref{fig:LTs:pxLT}, presents PX-LT for $\DSM^*$ (Table \ref{tab:Ex:DSfe1}) and individuals $\vec{x_1}$, $\vec{x_2}$. Nodes referring to the \textcolor{red}{\textbf{red}} and \textcolor{blue}{\textbf{blue}} masks are marked in the appropriate colors (the example in Section \ref{sec:relWork:PX}). 


\begin{thm}
    \label{thm:pxLTsAreCool}
\textcolor{black}{
For any problem with VIG $\VIG$, any perfect DSM $\DSM^*$ and individuals $\vec{x_{p1}}$ and $\vec{x_{p2}}$, the PX-LT creation algorithm will always yield an LT containing all PX masks as nodes.}
    \begin{proof}
    \textcolor{black}{
        We denote the DSM used for the clusterization as $\DSM^c$. It was obtained by the appropriate restriction of the set of indices, namely, we put $I=\{g:\vec{x_{p1}}[g] \neq \vec{x_{p2}}[g]\}$ and define $\DSM^c:I\times I \to[0,1]$ by $\DSM^c(g,h) = \DSM^*(g,h)$. All LT nodes will be joined in descending order starting from the highest value in $\DSM^c$. At each step a pair of genes $g$, $h$ with the largest entry in $\DSM^c$ (equivalently, $\DSM^*$) is chosen and the nodes containing $g$ and $h$ are merged into a new node. Let $\thr$ be a threshold for  $\DSM^*$, separating dependent pairs from the independent. After the last step before threshold crossing the following holds: (i) each gene $g$ was connected in one cluster with all genes $h$ for which there is a path $g_0$,$g_1$,...,$g_k$ with $g_0=g$, $g_k=h$ and $\DSM(g_i,g_{i+1})>\thr$, i.e., if $\VIG(g_i,g_{i+1})=1$; (ii) no gene $g$ was clustered with $h$ if $\DSM(g,h)<\thr$, i.e., if $\VIG(g,h)=0$. Hence LT contains all connected components of the graph $\VIG$ restricted to the vertices from $I$.}\par
    \end{proof}
\end{thm}


\subsection{PX-like Optimal Mixing}
\label{sec:pxLTs:pxOM}

Standard LTs support perturbation masks (LT nodes without the root) that are not dedicated to a given pair of individuals. Thus, they can be considered general, and the standard OM uses all (or almost all, OM in P3 ignores all masks of size $1$) LT-masks. PX-LT and the original PX support masks for a dedicated pair of individuals. However, even if PX-LT was built using $\DSM^*$, we do not know which masks are the PX masks. We may consider all PX-LT masks for each mixed individual pair, but the cost of such a mixing operator might be high. Therefore, we propose the PX-like Optimal Mixing operator (PX-OM) that utilizes PX-LT.\par

\begin{figure}
    \centering
	       \includegraphics[width=0.4 \linewidth]{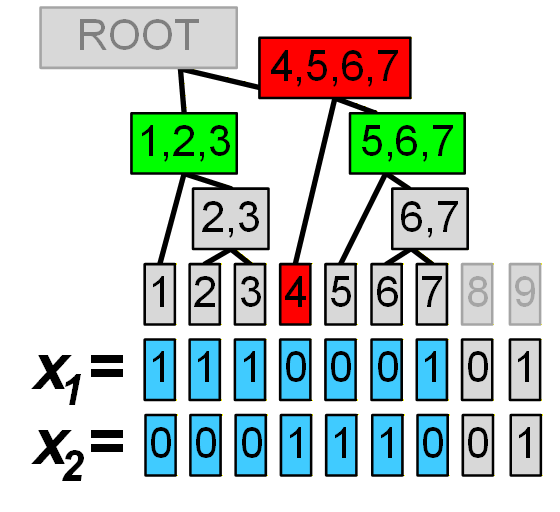}
	
	\caption{\textit{LTopWS} strategy for choosing PX-LT nodes}
	\label{fig:LTTop}
\end{figure}

\begin{algorithm}
    \small
	\caption{PX-like Optimal Mixing}
	\begin{algorithmic}[1]
        \Function{PX-OM}{$\DSM$, $\vec{x_{src}}$, $\vec{x_{don}}$}
            \State  $allMasks \gets $ ConstructLT($\DSM$, $\vec{x_{src}}$, $\vec{x_{don}}$);\label{alg:pxOM:LTnodesAll} \Comment{see Pseudocode \ref{alg:pxLTcreation}} 
            \State $maxMaskSize \gets $ Diff($\vec{x_{src}}$, $\vec{x_{don}}$) / 2;\label{alg:pxOM:LTnodesHalf}
            \State $masks \gets $ LTTopWithMaskSize($allMasks$, $maxMaskSize$);
            \State $masks \gets $ shuffle($masks$); \label{alg:pxOM:shuffle}
            \State $slideMs \gets $ empty;

            \For{\textbf{each} $mask$ \textbf{in} $masks$}
                \State $\vec{x'_{src}} \gets \vec{x_{src}}$; $\vec{x'_{src}} \xleftarrow[mask]{} \vec{x_{don}}$; \label{alg:pxOM:mixing}
                \If {$f(\vec{x'_{src}}) > f(\vec{x_{src}})$}
                    \Return{$\vec{x'_{src}}$}; \label{alg:pxOM:improvement}
                \EndIf
                \If {$f(\vec{x'_{src}}) = f(\vec{x_{src}})$}
                    $slideMs \gets slideMs + mask$; \label{alg:pxOM:slideSave}
                \EndIf
            \EndFor

            \If{$|slideMs| > 0$} \label{alg:pxOM:slideUseStart}
                \State $slideMask \gets $ GetRandomMask($slideMs$);
                \State $\vec{x'_{src}} \gets \vec{x_{src}}$; $\vec{x'_{src}} \xleftarrow[slideMask]{} \vec{x_{don}}$;
                \State \Return $\vec{x'_{src}}$; \label{alg:pxOM:slideUseEnd}
            \EndIf
            \State \Return $\vec{x_{src}}$; \label{alg:pxOM:noMod}
        \EndFunction
    \end{algorithmic}
	\label{alg:pxOM}
\end{algorithm}

In PX-OM, the entry information is the same as in the standard OM, i.e., $\DSM$, and the source and donor individuals ($\vec{x_{src}}$, $\vec{x_{don}}$, respectively). As a part of PX-OM (Pseudocode \ref{alg:pxOM}), we propose the \textit{LT Top With Size} (LTopWS) strategy for choosing masks from PX-LT. From the set of all PX-LT nodes (line \ref{alg:pxOM:LTnodesAll}), PX-OM chooses those that are of size larger than $1$ (not leaves) and not larger than half of the number of genes differing in $\vec{x_{src}}$, $\vec{x_{don}}$ (line \ref{alg:pxOM:LTnodesHalf}). The exact routine of node choosing is presented in Fig. \ref{fig:LTTop}. The \textcolor{blue}{\textbf{blue}} color marks all genes considered in PX-LT, the \textcolor{gray}{\textbf{gray}} color marks all genes that LTopWS did not analyze, and the \textcolor{red}{\textbf{red}} color marks nodes that were rejected due to their size. Finally, the \textcolor{darkgreen}{\textbf{green}} color marks the selected as masks. LTopWS starts from the nodes that are the root's children. If any of them has the appropriate size, it is accepted as the mixing mask and its subtree is not analyzed any further. In Fig. \ref{fig:LTTop}, node $\{x_1,x_2,x_3\}$ was accepted because its size is lower than $7/2=3.5$ and larger than 1. Thus, LTopWS did not consider the nodes in its subtree. Node $\{x_4,x_5,x_6,x_7\}$ was rejected because it is too large, and its children were considered. Node $\{x_4\}$ was rejected because it is a leaf, but node $\{x_5,x_6,x_7\}$ was accepted.\par

After selecting masks from PX-LT, their order is shuffled (line \ref{alg:pxOM:shuffle}), and each mask is used for mixing (line \ref{alg:pxOM:mixing}) until the improvement of $\vec{x_{src}}$ is found (line \ref{alg:pxOM:improvement}). The sliding masks that yield $\vec{x'_{src}}$ of fitness equal to $\vec{x_{src}}$ are stored in a separate list (line \ref{alg:pxOM:slideSave}). If no improvement is found, one sliding mask is randomly chosen and used to modify $\vec{x_{src}}$ (lines \ref{alg:pxOM:slideUseStart}-\ref{alg:pxOM:slideUseEnd}). If no improvement and no sliding masks were found, PX-OM returns unmodified $\vec{x_{src}}$ (line \ref{alg:pxOM:noMod}). Note that the LTopsWS strategy is similar to the LTop strategy proposed in \cite{wVIG}. \par

\subsection{PX-OM mixing masks for perfect DSM}
\label{sec:pxLTs:pxOMwithPerfect}

\newcommand{\dec}{\mathrm{dec}}

Let us perform a more detailed study of an example, in which we determine whether the created PX-like Linkage Trees contain all  PX masks. We define the standard deceptive function of order $k$ as 
\begin{equation}
    \label{eq:decStand}
    \mathit{dec}_k\bigl(u(\vec{x})\bigr)=
    \begin{cases}
        k - 1 - u & \text{if } u < k\\
        k & \text{if } u = k
    \end{cases}.
\end{equation}
Consider a six-gene problem $f(x_1,x_2,x_3,x_4,x_5,x_6)$, which is a sum of three standard deceptive functions, each two overlapping on one gene: $f(\vec{x}) = \dec_3(x_1,x_2,x_3)+\dec_3(x_3,x_4,x_5)+\dec_3(x_5,x_6,x_1)$. It is easy to show that any solution optimized by FIHC will end either as $000000$, $111111$, or one of a `hybrids': $111000$, $001110$ or $100011$.\\
We know by Theorem \ref{thm:pxLTsAreCool}, that to have all PX masks as nodes in PX-LT, all you need is a perfect DSM $\DSM^*$. To 
obtain one, we need the \emph{theoretical} DSM, i.e., the DSM based on theoretical distributions of gene pairs, to be perfect itself. In the given example we have four types of gene pairs: an overlapping gene with its neighbor (for example $x_1x_2$), two overlapping genes (e.g., $x_1x_3$), an overlapping and a non-overlapping genes from different domains (e.g., $x_1x_4$) and two non-overlapping genes (e.g. $x_2x_4$). The first two kinds are dependent pairs ($\VIG(x_1,x_2) = \VIG(x_1,x_3) = 1$) and the other two are independent. To find the theoretical distributions of these pairs, one must count the probabilities of obtaining each of possible outcomes of the FIHC optimization. Let $p_0$, $p_1$ and $p_h$ denote the probabilities that a randomly chosen solution will end as $000000$, $111111$ and a a fixed hybrid (say $111000$), respectively. Clearly, $p_0+p_1+3p_h=1$, and any probability of a pair of genes can be derived from these values, e.g., $P(x_1x_2=00)=p_0+p_h$.
The calculation of $p_0$, $p_1$ or $p_h$ turns out to be tedious, 
nevertheless, we calculated the probabilities, and the entries of the theoretical DSM are approximately: $D(x_1,x_2) = 0.47$, $D(x_1,x_3) = 0.19$, $D(x_2,x_4) = 0.17$ and $D(x_1,x_4) = 0.08$ (see Section S-II, supplementary material, for details and further discussion). These numbers indeed  give a perfect DSM. By the Law of Large Numbers,  with high probability, a sufficiently large population will provide a DSM with entries close to theoretical, making it also a perfect DSM. Then, PX-LTs \textbf{will contain} all PX masks. However, the dependency strengths of pairs $(x_1,x_3)$ and $(x_2,x_4)$ are close, so obtaining a perfect DSM empirically may require large populations.
By the full search we have also found probabilities for the sum of three $\dec_4$ functions, overlapping at single genes. The theoretical DSM entries are approximately: $D(x_1,x_2) = 0.34$, $D(x_1,x_4) = 0.15$, $D(x_2,x_3) = 1.0$ (representatives of all types of dependent pairs), $D(x_1,x_5) = 0.02$, $D(x_2,x_5) = 0.02$ (representatives of independent pairs). Note that as we enlarge the domains of the subfunctions, the dependence of a pair of two overlapping genes from a common domain is still the weakest, but it differs more from the values of independent genes.\par

A more detailed analysis of the size of a population sufficient for obtaining a perfect DSM in the spirit of \cite{sllForBimodals}, is an interesting research direction. As for now, we remark that 
it is impossible to statistically detect any dependencies in the absence of `hybrids'. 
Knowing $p_h$, by an inclusion-exclusion principle we get that with probability $0.99$ FIHC will introduce a single fixed `hybrid' into a random population of 50 solutions, two fixed `hybrids' in a population of 57 solutions, and all of them within 62 solutions.

\section{Noise introduction and its consequences for the non-monotonical dependencies}
\label{sec:noiseMonotonicity}

Section \ref{sec:relWork:nooise} points out that the real-world problem instances may be the subject of noise and that, in the general case, the function that directly transforms solution encoding into solution evaluation may be unknown. Therefore, linkage learning optimizers that can solve such instances effectively and efficiently are needed. Mechanisms introducing noise into the well-known benchmarks are useful tools in obtaining this objective. The intuitions behind introducing the noise into the considered problem are as follows \cite{wVIG}.

\textbf{I1.} Noise influence should be negligible in terms of problem difficulty. Here, we interpret it that the location of the high-quality locally optimal solutions should be unaffected by the noise.\par
\textbf{I2.} For highly effective linkage learning optimizers, the main difficulty introduced by noise is the appearance of variable dependencies it creates, which do not help solve the considered instance.\par

To simulate such noise, one can add low-valued functions taking two randomly chosen arguments \cite{wVIG}, i.e., $f_{noised}(\vec{x}) = f_{true}(\vec{x}) + f_{noise}(\vec{x})$, where $f_{true}(\vec{x})$ is the function into which we introduce noise and $f_{noise}(\vec{x}) = \sum_{i=1}^{N} f_i(x_{i,1}, x_{i,2})$, where $f_i$ are the low-valued functions ($f_{noise}(\vec{x}) << f_{true}(\vec{x})$ for any $\vec{x}$) simulating noise, $N$ is the number of such functions, and $x_{i,1}$ and $x_{i,2}$ are the arguments randomly chosen from $\{x_1,...,x_n\}$. Note that noise representation using subfunctions with two variables is coherent with the noise characteristics presented in \cite{wVIG}.\par

Consider a function $f_{e5}(x_1,...,x_{8}) = f_{e5,true}(\vec{x}) + f_{e5,noise}(\vec{x})$ such that $f_{e5,true}(\vec{x}) = dec_4(x_1,...x_4) + dec_4(x_5,...x_8)$ and $f_{e5,noise}(\vec{x}) = \sum_{i=1}^{N} f_i(x_{i,1}, x_{i,2})$, where $\sum_{i=1}^{N} f_i(x_{i,1}, x_{i,2}) < 1$ for any $\vec{x} = [x_1,...,x_8]$. $f_{e5,noise}$ can raise non-linear dependencies in $f_{e5}$ for $\{x_{i,1}, x_{i,2}\}$ pairs. However, the monotonic relation of any two solutions, say $\vec{x_a}$, $\vec{x_b}$, such that $f_{e5}(\vec{x_a}) < f_{e5}(\vec{x_b})$ or $f_{e5}(\vec{x_a}) > f_{e5}(\vec{x_b})$, will not be affected by $f_{e5,noise}$. Indeed, $f_{e5,noise}$ can only raise non-monotonical dependencies when $f_{e5,true}(\vec{x_a}) = f_{e5,true}(\vec{x_b})$. Thus, the non-mono\-to\-nicity check is much less sensitive to noise and is more suitable to handle noised instances than the non-linearity check \cite{wVIG}.\par

The advantage of the non-monotonicity check presented above makes it challenging to introduce noise characterized by features I1 and I2 into benchmark instances. Consider function $f_{e6}$ defined as
\begin{equation}
    \label{eq:nonMonoNoiseExample}
    f_{e6}(x_1,...,x_8)=
    \begin{cases}
        5.5 & \!\!\text{if } \vec{x} = [0110\,0000]\\
        dec_4(x_1,...x_4) + dec_4(x_5,...x_8) & \!\!\text{otherwise}
    \end{cases}
\end{equation}
Below, we will refer to $\vec{x_{c1}} = [x_1,...,x_4]$ and $\vec{x_{c2}} = [x_5,...,x_8]$, i.e., $\vec{x} = [\vec{x_{c1}}, \vec{x_{c2}}]$. Note that all variables in $\vec{x_{c1}}$ and in $\vec{x_{c2}}$ are non-monotonically dependent on each other. In $f_{e5}$, that is similar to $f_{e6}$, there are no non-monotonical dependencies joining variables from $\vec{x_{c1}}$ and $\vec{x_{c2}}$. However, for $f_{e6}$, the situation differs.\par

Consider $\vec{x_{c2,0}} = [0000]$ and $\neg \vec{x_{c2,0}} \neq [0000]$. For $\vec{x_{c2,0}}$, it is true that $f_{e6}(0110, \vec{x_{c2,0}}) > f_{e6}(0010, \vec{x_{c2,0}}) = f_{e6}(0100, \vec{x_{c2,0}})$. However, for any value of $\neg \vec{x_{c2,0}}$, these relations are reverted, i.e., $f_{e6}(0110, \neg\vec{x_{c2,0}}) < f_{e6}(0010, \neg\vec{x_{c2,0}}) = f_{e6}(0100, \neg\vec{x_{c2,0}})$. Thus, flipping any variable in $\vec{x_{c2,0}}$ changes the monotonic relations between solution pairs \{[0110 ****], [0010 ****]\} and \{[0110 ****], [0100 ****]\}, which leads to non-monotonical dependency between variables $x_2$ and $x_3$ and each of the variables $x_5,...,x_8$.\par

Consider $f_{e6}$ as $f_{e6, true}(x_1,...,x_8) = dec_4(x_1,...x_4) + dec_4(x_5,...x_8)$ with noise, where the value $f_{e6, true}(0110\text{ }0000) = 4$ is noised and equals $5.5$ instead. The four local optima of $f_{e6, true}$, i.e., $[0000\text{ }0000]$, $[0000\text{ }1111]$, $[1111\text{ }0000]$, and $[1111\text{ }1111]$, remain valid for $f_{e6}$. Thus, the $f_{e6}$ example shows that it is possible to introduce noise that raises non-monotonical dependencies between variables and preserves I1, I2 features, i.e., the local optima of the original function ($f_{e6,true}$) are not changed, and finding the optimum requires processing all variables in each of the $\vec{x_{c1}}$ and $\vec{x_{c2}}$ jointly.\par

Therefore, we propose the following mechanism of noise introduction into benchmarks. For $f_{noised}(\vec{x}) = f_{true}(\vec{x}) + f_{noise}(\vec{x})$, where $f_{true}(\vec{x})$ is the original function, we define $f_{noise}(\vec{x})$ as
\begin{equation}
    \label{eq:noiseProposition}
    f_{noise}(\vec{x}) =  
    \begin{cases}
        V(\vec{x}, L_{noise}) & \text{if } u(\vec{x_{noise}}) \mod F_{noise} = 0\\
        0 & \text{otherwise}
    \end{cases}
\end{equation}
where $\vec{x_{noise}}$ is a randomly chosen subset of $\vec{x}$ such that $N_{size} = |\vec{x_{noise}}|$, $N_{size}$, $L_{noise}$, and $F_{noise}$ are user-defined parameters and $V(\vec{x}, L_{noise})$ is defined as
\begin{equation}
    \label{eq:noiseVolume}
    V(\vec{x}, L_{noise}) =  
    \begin{cases}
        0 & \text{if } f_{true}(\vec{x}) \geq L_{noise}\\
        L_{noise} - f_{true}(\vec{x}) & \text{otherwise}
    \end{cases}
\end{equation}
The above proposition is convenient. $N_{size}$ defines how severe will be the noise, the higher the $N_{size}$ is, the more non-monotonical dependencies will originate from the noise. The higher $f_{true}(\vec{x})$ is, the lower will be the influence of noise and for the solutions of the highest quality, i.e., $f_{true}(\vec{x}) \geq L_{noise}$, then $f_{noised}(\vec{x}) = f_{true}(\vec{x})$, which shall preserve the difficulty of solving $f_{true}(\vec{x})$.

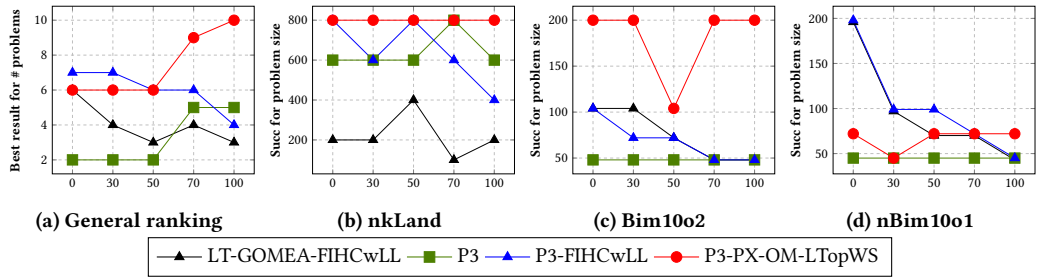
\begin{figure*}[]
    \begin{subfigure}[b]{0.19\linewidth}
		\resizebox{\linewidth}{!}{%
			\tikzset{every mark/.append style={scale=2.5}}
			\begin{tikzpicture}
			\begin{axis}[%
            legend entries={LT-GOMEA-FIHCwLL, P3, P3-FIHCwLL, P3-PX-OM_LTopWS},
            legend columns=-1,
            legend to name=named,
			xtick={1,2,3,4,5},
			xticklabels={0,30,50,70,100},
			xmin=0.5,
			xmax=5.5,
			ylabel=\textbf{Best result for \# problems},
			grid,
			grid style=dashed,
			ticklabel style={scale=1.5},
			label style={scale=1.5},
			legend style={font=\fontsize{8}{0}\selectfont}
			]

             \addplot[
			color=black,
			mark=triangle*,
			]
			coordinates {
				(1,6)(2,4)(3,3)(4,4)(5,3)
			};
   
            \addplot[
			color=darkgreen,
			mark=square*,
			]
			coordinates {
				(1,2)(2,2)(3,2)(4,5)(5,5)
			};

            \addplot[
			color=blue,
			mark=triangle*,
			]
			coordinates {
				(1,7)(2,7)(3,6)(4,6)(5,4)
			};
   
			\addplot[
			color=red,
			mark=*,
			]
			coordinates {
				(1,6)(2,6)(3,6)(4,9)(5,10)
			};

			\end{axis}
			\end{tikzpicture}
		}
		\caption{General ranking}
		\label{fig:scalab:general}
	\end{subfigure}
    \begin{subfigure}[b]{0.19\linewidth}
		\resizebox{\linewidth}{!}{%
			\tikzset{every mark/.append style={scale=2.5}}
			\begin{tikzpicture}
			\begin{axis}[%
            legend entries={LT-GOMEA-FIHCwLL, P3, P3-FIHCwLL, P3-PX-OM_LTopWS},
            legend columns=-1,
            legend to name=named,
			xtick={1,2,3,4,5},
			xticklabels={0,30,50,70,100},
			xmin=0.5,
			xmax=5.5,
			ylabel=\textbf{Succ for problem size},
			grid,
			grid style=dashed,
			ticklabel style={scale=1.5},
			label style={scale=1.5},
			legend style={font=\fontsize{8}{0}\selectfont}
			]

            \addplot[
			color=black,
			mark=triangle*,
			]
			coordinates {
				(1,200)(2,200)(3,400)(4,100)(5,200)
			};
   
            \addplot[
			color=darkgreen,
			mark=square*,
			]
			coordinates {
				(1,600)(2,600)(3,600)(4,800)(5,600)
			};

            \addplot[
			color=blue,
			mark=triangle*,
			]
			coordinates {
				(1,800)(2,600)(3,800)(4,600)(5,400)
			};
   
			\addplot[
			color=red,
			mark=*,
			]
			coordinates {
				(1,800)(2,800)(3,800)(4,800)(5,800)
			};

			\end{axis}
			\end{tikzpicture}
		}
		\caption{\textbf{nkLand}}
		\label{fig:scalab:nkLand}
	\end{subfigure}
    \begin{subfigure}[b]{0.19\linewidth}
		\resizebox{\linewidth}{!}{%
			\tikzset{every mark/.append style={scale=2.5}}
			\begin{tikzpicture}
			\begin{axis}[%
            legend entries={LT-GOMEA-FIHCwLL, P3, P3-FIHCwLL, P3-PX-OM_LTopWS},
            legend columns=-1,
            legend to name=named,
			xtick={1,2,3,4,5},
			xticklabels={0,30,50,70,100},
			xmin=0.5,
			xmax=5.5,
			ylabel=\textbf{Succ for problem size},
			grid,
			grid style=dashed,
			ticklabel style={scale=1.5},
			label style={scale=1.5},
			legend style={font=\fontsize{8}{0}\selectfont}
			]

            \addplot[
			color=black,
			mark=triangle*,
			]
			coordinates {
				(1,104)(2,104)(3,72)(4,48)(5,48)
			};
   
            \addplot[
			color=darkgreen,
			mark=square*,
			]
			coordinates {
				(1,48)(2,48)(3,48)(4,48)(5,48)
			};

            \addplot[
			color=blue,
			mark=triangle*,
			]
			coordinates {
				(1,104)(2,72)(3,72)(4,48)(5,48)

			};
   
			\addplot[
			color=red,
			mark=*,
			]
			coordinates {
				(1,200)(2,200)(3,104)(4,200)(5,200)
			};
            
			\end{axis}
			\end{tikzpicture}
		}
		\caption{\textbf{Bim10o2}}
		\label{fig:scalab:Bim10o2}
	\end{subfigure} 
    \begin{subfigure}[b]{0.19\linewidth}
		\resizebox{\linewidth}{!}{%
			\tikzset{every mark/.append style={scale=2.5}}
			\begin{tikzpicture}
			\begin{axis}[%
            legend entries={LT-GOMEA-FIHCwLL, P3, P3-FIHCwLL,P3-PX-OM-LTopWS},
            legend columns=-1,
            legend to name=named,
			xtick={1,2,3,4,5},
			xticklabels={0,30,50,70,100},
			xmin=0.5,
			xmax=5.5,
			ylabel=\textbf{Succ for problem size},
			grid,
			grid style=dashed,
			ticklabel style={scale=1.5},
			label style={scale=1.5},
			legend style={font=\fontsize{8}{0}\selectfont}
			]

            \addplot[
			color=black,
			mark=triangle*,
			]
			coordinates {
                (1,196)(2,97)(3,70)(4,70)(5,43)
			};
   
            \addplot[
			color=darkgreen,
			mark=square*,
			]
			coordinates {
				(1,45)(2,45)(3,45)(4,45)(5,45)

			};

            \addplot[
			color=blue,
			mark=triangle*,
			]
			coordinates {
				(1,198)(2,99)(3,99)(4,72)(5,45)
			};
   
			\addplot[
			color=red,
			mark=*,
			]
			coordinates {
				(1,72)(2,45)(3,72)(4,72)(5,72)
			};
            
			\end{axis}
			\end{tikzpicture}
		}
		\caption{\textbf{nBim10o1}}
		\label{fig:scalab:nBim10o1}
	\end{subfigure}
	
	\pgfplotslegendfromname{named}

	\caption{Ranking scalability. X-axis represents $N_{size}$ as a percentage of the total length of the genotype, i.e., for 100\%, $N_{size} = n$}
	\label{fig:scalab}
\end{figure*}

\section{Experiments}
\label{sec:expMain}

\subsection{Experiment setup and considered optimizers}
\label{sec:expMain:setup}
The objective of the experiments is to verify if PX-OM-LTopWS can increase the effectiveness of the SLL-using optimizers when applied to solve noised problems with hidden structure. Therefore, we introduce PX-OM-LTopWS into the original P3 and LT-GOMEA. We may expect that for the \textit{non}-noised problems, SLL-using optimizers will be outperformed by their FIHCwLL-using versions. However, with the increase of the noise range ($N_{size}$), we may expect that eVIGs maintained by the FIHCwLL-using optimizers will become full graphs. Then, their effectiveness should become similar to their original versions because using FIHCwLL will become irrelevant.\par

To emulate noise we define $N_{size}$ as a peercentage of $n$. We consider $N_{size}$ values of 0\% (non-noised problems), 30\%, 50\%, 70\%, and 100\%. The considered problem set is typical for optimizers dedicated to combinatorial optimization and concerns NK-landscapes \cite{nkLandscapes}, Ising Spin Glasses~\cite{isg}, max3sat~\cite{P3Original,transTokBounded}, standard, bimodal and noised bimodal deceptive functions concatenations and cyclic traps \cite{cyclicTrap}. Bimodal, noised, and standard deceptive functions concatenations are denoted as \textit{Bim10}, \textit{nBim10}, and \textit{Dec5}, respectively. The names of the cyclic traps based on these functions are suffixed with a number indicating how many variables were shared between the two neighbouring functions. For instance, \textit{Bim10o2} is the concatenation of bimodal functions of order 10, where each bimodal function shares 2 variables with each of its neighbours. \par

All problems and optimizers were joined in one C++ project available at GitHub\footnote{\url{https://github.com/przewooz/PXwithDSM}} and shared as many source code parts as possible. The stop condition was set to $20\cdot 10^7$ FFE.\par

To compare the effectiveness subject to the increasing noise, for each problem, we considered the instance size with at least 80\% success rate. Since the maximum solved instance size frequently differed, such an experimental protocol remains simple and does not require statistical test confirmation (if one optimizer can solve instances of a larger size it outperforms the other).

\subsection{SLL-using optimizer versions comparison}
\label{sec:expMain:results}

\begin{table}[]
\caption{P3-based SLL-using optimizers comparison} 
    \label{tab:p3:versions}
\scriptsize

\begin{tabular}{l|rrr|rrr|rrr}

&      \multicolumn{3}{c}{\textbf{P3-PX-OM-LTopWS}}    & \multicolumn{3}{c}{\textbf{P3-PX-OM-All}}    & \multicolumn{3}{c}{\textbf{P3}}               \\
         
         &      & \textbf{Opt}         & \textbf{FFE} &      & \textbf{Opt}         & \textbf{FFE}&      & \textbf{Opt}         & \textbf{FFE}\\

         & \textbf{Size} & \textbf{[\%]}             & \textbf{Med}                   & \textbf{Size} & \textbf{[\%]}             & \textbf{Med}& \textbf{Size} & \textbf{[\%]}             & \textbf{Med}\\
         \hline

\textbf{nkLand}   & \textbf{800}  & 100             & 6.6E+6                & 200  & 100                   & 4.7E+6                      & 600  & 80      & 8.6E+6         \\
\textbf{ISG}      & \textbf{841}  & 100             & 5.1E+5                & \textbf{841}  & 97                    & 4.9E+6                      & \textbf{841}  & 100     & 5.7E+5         \\
\textbf{m3s}      & 75   & 80              & 9.0E+4                & 50   & 93                    & 3.1E+4                      & \textbf{100}  & 93      & 2.0E+5         \\
\textbf{Bim10}    & \textbf{200}  & 93              & 3.6E+6                & 100  & 90                    & 9.8E+6                      & 80   & 100     & 5.0E+6         \\
\textbf{Bim10o1}  & \textbf{198}  & 90              & 7.4E+6                & 72   & 80                    & 9.0E+6                      & 45   & 100     & 1.8E+6         \\
\textbf{Bim10o2}  & \textbf{200}  & 97              & 8.6E+6                & 48   & 97                    & 5.1E+6                      & 48   & 100     & 8.0E+6         \\
\textbf{Bim10o3}  & \textbf{98}   & 100             & \textbf{4.3E+6}               & 98   & 100                   & 1.4E+7                      & 28   & 100     & 3.9E+6         \\
\textbf{nBim10}   & \textbf{200}  & 93              & 3.6E+6                & 100  & 90                    & 9.8E+6                      & 80   & 100     & 5.0E+6         \\
\textbf{nBim10o1} & \textbf{80}   & 100             & \textbf{1.1E+7}                & 50   & 100                   & 2.9E+6                      & \textbf{80}   & 97      & \textbf{9.7E+6}         \\
\textbf{nBim10o2} & 7\textbf{}2   & 100             & 1.4E+7                & 45   & 100                   & 3.8E+6                      & 45   & 100     & 1.9E+6         \\
\textbf{nBim10o3} & \textbf{49}   & 100             & 6.1E+6                & 49   & 80                    & 1.5E+7                      & 49   & 100     & \textbf{4.8E+6}         \\
\textbf{Dec5}   & 200  & 100             & \textbf{4.8E+4}                & 200  & 100                   & 3.4E+5                      & 200  & 100     & \textbf{3.2E+4}         \\
\textbf{Dec5o1}   & 200  & 100             & \textbf{5.8E+4}               & 200  & 100                   & 3.9E+5                      & 200  & 100     & \textbf{6.6E+4}         \\
\textbf{Dec5o2}   & 201  & 100             & \textbf{1.1E+5}                & 201  & 100                   & 5.9E+5                      & 201  & 100     & \textbf{8.4E+4} \\
\hline
\textbf{Most eff.} & \multicolumn{3}{c}{\textbf{13}}    & \multicolumn{3}{c}{\textbf{1}}    & \multicolumn{3}{c}{\textbf{7}}            
\end{tabular}
\end{table}

Table \ref{tab:p3:versions} presents P3--based optimizers comparison for non-noised instances. For each optimizer, we consider its original version, version with PX-OM using LTopWS, and the version considering all masks from PX-LT. The results indicate that for the LTopWS strategy performs significantly better. Table S-I (suuplementary material) presents the information for LT-GOMEA--based optimizers, which shows that PX-OM is beneficial for P3 but not for LT-GOMEA. Therefore, in the latter part of experiments, we consider only one optimizer using PX-OM, namely P3-PX-OM-LTopWS.


\begin{table}[]
\caption{The median of the average percentage of PX masks in PX-LT trees for various problems in the PX-OM-LTopWS runs ($N_{size}$ equal to 100\% of the problem size).} 
    \label{tab:pxMasksPerc}
\scriptsize
\begin{tabular}{l|rr|l|rr|l|rr}
         & \textbf{Size} &  \textbf{PX[\%]} & & \textbf{Size} &  \textbf{PX[\%]}& & \textbf{Size} &  \textbf{PX[\%]}      \\
         \hline
\textbf{nkL}   & 400                 & 13.84 & \textbf{Bim10}    & 100        & 10.96  & \textbf{nBim10}   & 80      & 1.89  \\
\textbf{ISG}      & 400                 & 53.30 & \textbf{Bim10o1}  & 72         & 8.29  & \textbf{nBim10o1} & 45       & 0.96  \\
\textbf{m3s}      & 75                  & 32.29 & \textbf{Bim10o2}  & 72         & 7.73  & \textbf{nBim10o2} & 48       & 1.15  \\
 \textbf{Dec5}     & 200   & 37.33    & \textbf{Bim10o3}  & 49         & 8.13  & \textbf{nBim10o3} & 49       & 2.26  
  
\end{tabular}
\end{table}

Table \ref{tab:pxMasksPerc} presents the percentage of PX masks covered by the PX-LTs nodes during the whole run. The percentage was similar regardless of the noise and can be considered high. The same analysis performed for other optimizers (we considered the exchanged variables during OM) showed that the percentage of the PX-like mixing masks was below 1\% for all considered instances. Thus, the obtained percentage of the PX-like masks can be considered high.\par


\subsection{Main results}
\label{sec:expMain:results}

Fig. \ref{fig:scalab} presents the general results in the graphical form. All results concerning standard deceptive functions (i.e., Dec5, Dec5o1, and Dec5o2) were excluded from this part because the differences between the considered optimizers were negligible for these problems. The general ranking presenting the number of problems for which a given optimizer solved the instances of the largest size is given in Fig. \ref{fig:scalab:general}. The effectiveness of P3-FIHCwLL and LT-GOMEA-FIHCwLL is high for problems without noise. However, the effectiveness of P3-PX-OM-LTopWS is relatively high and significantly higher than P3. As the noise range, i.e., $N_{size}$, increases, the relative effeectiveness of the PX-OM-using optimizer increases and the effectiveness of P3-FIHCwLL becomes similar to P3. The same phenomenon can be observed in Fig.\ref{fig:scalab:nkLand}-\ref{fig:scalab:nBim10o1} presenting the scalability for various problems. For the NK-landscapes and Bim10o2, P3-PX-OM-LTopWS is effective regardless the noise. However, for nBim10o1, the advantage of FIHCwLL-using P3 and LT-GOMEA over both SLL-using optimizers is significant. However, it disappears when the noise range raises, which does not affect P3-PX-OM-LTopWS.

\section{Conclusions}
\label{sec:conc}

In this work, we have proposed a new DSM clusterization algorithm that aims at constructing such LTs that some of their nodes are equal to PX mixing masks. The proposed procedure is proven to yield all PX mixing masks for a given pair of individuals if the LT is built on the basis of a perfect DSM, a new concept that we introduce to justify our proposal. The obtained results show that PX-OM, the proposed mixing operator, significantly improves the effectiveness of P3 and remains unaffected by the increasing noise. \par 

Our proposition may have a significant influence on the real-world optimization scenarios that face noisy instances. For such problems, many of the state-of-the-art optimizers may become ineffective. Finally, the presented analysis enables new research directions, i.e., the further analysis of the probability of obtaining a perfect or near-perfect DSM for various problems and the potential that lies in proposing new algorithms for the DSM clusterization.

\begin{acks}
This work was supported by: Polish National Science Centre (NCN), 2022/45/B/ST6/04150 (B. Frej, M.M. Komarnicki, M.Prusik, M.W. Przewozniczek); FAPESP, \#2024/15430-5 and \#2024/08485-8, CNPq, \#304640/2024-7 (R. Tin\'os).
\end{acks}
	
	\bibliographystyle{ACM-Reference-Format}
	\bibliography{PXwithDSM} 
	
\end{document}